# A Conditional Generative Chatbot using Transformer Model

**Nura Esfandiari [a], Kourosh Kiani [a*], Razieh Rastgoo [a]**
[a] Electrical and Computer Engineering Faculty, Semnan University, Semnan, 3513119111, Iran
[*] Corresponding author (Kourosh.kiani@semnan.ac.ir)
Email: nura_esfandiari@semnan.ac.ir, Kourosh.kiani@semnan.ac.ir, rrastgoo@semnan.ac.ir

A B S T R A C T

A Chatbot serves as a communication tool between a human user and a machine to achieve an appropriate answer based on the human input. In more recent approaches, a combination of Natural Language Processing and sequential models are used to build a generative Chatbot. The main challenge of these models is their sequential nature, which leads to less accurate results. To tackle this challenge, in this paper, a novel architecture is proposed using conditional Wasserstein Generative Adversarial Networks and a transformer model for answer generation in Chatbots. While the generator of the proposed model consists of a full transformer model to generate an answer, the discriminator includes only the encoder part of a transformer model followed by a classifier. To the best of our knowledge, this is the first time that a generative Chatbot is proposed using the embedded transformer in both generator and discriminator models. Relying on the parallel computing of the transformer model, the results of the proposed model on the Cornell Movie-Dialog corpus and the Chit-Chat datasets confirm the superiority of the proposed model compared to state-of-the-art alternatives using different evaluation metrics.

## 1. Introduction

Regarding the increasing use of social networks, a new technology called Chatbot has been developed as a tool for human-computer interaction. Chatbots have been widely applied in various fields such as e-commerce [1, 2], education [3], banking and insurance [4], Data collection and management [5], and Health [6]. Chatbots automatically give more attractive answers to users through easy and efficient communications. The key factor in the suitable design of Chatbots is to provide understandable answers to the user [7]. To this end, various approaches have been recently developed to build Chatbots. In general, the approaches of Chatbot development can be classified into two categories: open and closed domain (See Fig. 1). Chatbots with the ability to answer on more than one domain, are called open domains. In contrast, closed domain Chatbots can answer only to questions concerning a particular domain.

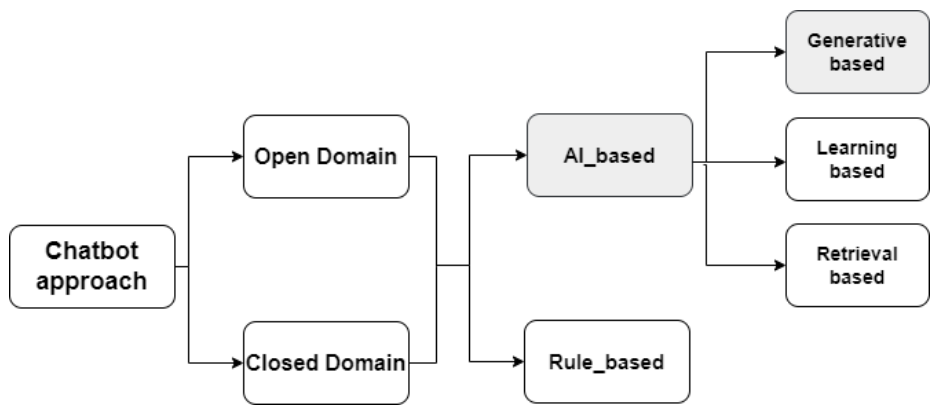

Fig. 1. Classification of Chatbot approach

Open and closed domains Chatbots can be categorized into Rule-based and Artificial Intelligence (AI)-based Chatbots. Rule-based Chatbots rely on the user's input with a base template. This type of Chatbots choose a predefined answer from a set of answers [8]. However, they are inflexible and limited to some predefined answers. ELIZA and ALICE are based on this approach [9]. AI-based Chatbots are trained to have human-like conversations using a hybrid of NLP and deep learning approaches. More concretely, AI-based Chatbot classified into three categories: Retrieval, Learning, and Generative-based methods. The first category, the retrieval-based method, chooses the most similar answer from a dataset using a functional scoring metric [10]. The second category, the learning-based method, usually learns patterns from training data, containing questions with answers, through deep learning models to generate relevant answers. The last category, the generative-based method, is usually based on the Sequence to Sequence (Seq2Seq) learning models [11-14]. However, the main challenge of these models is their sequential nature, which is led to less accurate results. To tackle this challenge, in recent years, researchers have developed some models based on the transformer models [15-19]. Relying on the parallel computing as well as the self-attention mechanism of the transformer model, the performance of the models developed using transformers has been improved in comparison with the Seq2Seq models [20]. Although, Learning-based models are not able to generate various answers. To overcome this shortcoming, the generative-based models have been developed to learn the answers distribution.

Some models such as Sequence Generative Adversarial Networks (SeqGAN) [21] and stepwise GAN (StepGAN) [22] have been developed as the third category, the generative-based model. The sequential nature of these models is led to use reinforcement learning techniques for completing the generator's answers. These models suffer from slow convergence due to their high variance and low processing speed. To tackle the challenge of sequential nature as well as the accuracy improvement in these models, a novel architecture is proposed based on conditional Wasserstein Generative Adversarial Networks (cWGAN) using transformer model which processes parallelly during the training phase. This architecture enhances the efficiency through generated human-like answers.

The main contributions of this paper can be listed as follows: 1) Model: We propose a novel model using the transformer model, which is used in both generator and discriminator of a cWGAN. To the best of our knowledge, this is the first time that such a model is proposed in Chatbot. 2) To tackle less accurate results in previous seq2seq model, we using the cWGAN and a transformer model for answer generation in Chatbots. 3) Extensive experiments conducted on two challenging datasets show that our architecture outperforms state-of- the-art methods in the field.

The rest of this paper is organized as follows: Section 2 reviews related works in Chatbot. The proposed architecture is explained in Section 3. In the following, the experimental results and discussion are presented in Section 4 and 5, respectively. Finally, conclusion and future works are provided in Section 6.

## 2. Literature review

In this section, we briefly review the recent works in four categories: Rule-based, Retrieval -based, Learning-based, and Generative-based.

### 2.1. Rule-based models

In the rule-based models, the characteristic variables of the input expression are first specified. Then, a predefined answer is provided based on the variables and rules [7]. Rule-based approaches can be divided into two categories: pattern matching methods and standard task-oriented systems [23]. In pattern matching methods, Chatbots match the user's input to the pattern of the rules and select a predetermined answer from the set of answers using pattern matching algorithms. Task-oriented systems guide the user to complete certain items. Since the 1990s, a great deal of research has been conducted into the design of Chatbots based on similar rules for providing services in specific domains. These Chatbots are known as task-oriented modular chat systems that guide the user to perform some structured tasks, such as restaurant and film reservations. Since 1966, the development of the Eliza Chatbot has begun with a pattern-based approach. This Chatbot analyzed the input sentence based on the parsing rules established by the keywords in a sentence [24]. "Pari" adds some influential variables like "fear", "anger" and "distrust" to the more complex rules. These rules have made the conversation more humane-like [25]. ALICE uses Artificial Intelligence Markup Language (AIML), which is the category constituting the unit of knowledge to combine a template and an optional field [26]. In addition, several platforms, such as Microsoft LUIS, IBM Watson Assistant, and Dialog flow, have been developed to assist the user in making Chatbots. These types of Chatbots have drawbacks despite their simplicity, quick implementation, and cost-effectiveness. Inflexibility, lack of learning, inability to create new answers, and being limited to some predefined answers are the most important challenges of rule-based Chatbots.

### 2.2. Retrieval-based

The retrieval-based Chatbots select the best matching answer for the user's question by searching a pre-constructed conversational repository [27]. Lowe et al. [28] developed a retrieval-based Chatbot using the Term Frequency - Inverse Document Frequency (TF-IDF) method. The TF-IDF vectors of each candidate's question and answer are computed by concatenating all TF-IDF scores. The candidate answers with the highest cosine similarity to the question vector are selected as the final answers. Lu et al. [29] proposed an architecture to overcome the short-text matching problem of the developed models. Later, Convolutional Neural Network (CNN), Recurrent Neural Network (RNN) and its extensions such as Long Short-Term Memory (LSTM) and Gated Recurrent Unit (GRU) were widely used in this field [23]. For instance, Zhou et al. proposed an approach by adding an attention mechanism to the deep network to provide question and answer matching [30]. In another work, Shu et al. [31] proposed a retrieval-based model to search for answers related to the question by combining keyword extraction modules and two-stage transformer. In overall, while the retrieval-based approach is widely used by researchers, no new answer is generated and only the most probable answer is retrieved from the database. This can restrict the Chatbot answers. Moreover, it is essential to have a database in the inference phase.

### 2.3. Learning-based

This approach is usually based on Seq2Seq learning model. In the case of long sentences and conversations (more than 20 words), all essential information of a source sentence should be compressed into a fixed-length vector which is challenging [20]. To tackle this challenge, different approaches have been suggested by researchers. For instance, making the structural changes, such as adding the word embedding matrix [32], modification of the encoder or decoder [33, 34], and adding the attention mechanism [11, 13, 14, 35] are some of the solutions suggested by researchers. In addition to this, some researchers have used transformer-based model, that replaces the recurrent layers commonly used in encoder-decoder architectures with multi-headed self-attention. Rao et al. [36], have developed a hybrid model for deep transfer learning-based text generation using the Elmo language model for embedding, Variational Autoencoder (VAE), and Bi-directional Long Short-Term Memory (BiLSTM). A transformer-based answer generation model, named DIALOGPT, was also presented as a pretrained model by Zhang et al. [37]. This model is publicly released to facilitate the development of more intelligent dialogue systems. Another suggested model is a Chatbot model using a Bidirectional Encoder Representations from Transformers (BERT) model, which only has an encoder [38]. Lin et al. [39] introduce CAiRE, an end-to-end empathy conversation agent. The proposed model leverages a large-scale pre-trained language model and fine-tunes it using multiple objectives, such as response language modeling, response prediction, and dialogue emotion detection. Also, Nuruzzaman and Hussain[40] suggest a domain-specific chatbot namely IntelliBot, which is a dialogue-based chatbot using multiple strategies (rule-based, retrieval-based, and learning-based) to generate a response. While the learning-based models have obtained promising results, they cannot generate various answers due to not having the ability to learn the answer distribution.

### 2.4. Generative-based

Generative-based methods learn the answer distribution using the generative models, such as SeqGAN [21] and StepGAN [22]. These models use the Reinforcement Learning (RL) techniques. In the SeqGAN, The RL reward signal comes from the discriminator, which is judged on a complete sequence, and fed back to the intermediate state action steps using Monte Carlo search. However, the Monte Carlo Tree Search (MCTS) used in SeqGAN has a high computational efficiency. In the StepGAN approaches, the Generative adversarial network (GAN) is evaluated in a step-wise manner. In this way, the discriminator is modified to automatically assign scores and determine the suitability of each generated subsequence. Wu and Wang added a new loss function (Truth guidance) to achieve a closer generated text to the real data [41]. Also, a discriminator network model was designed based on the self-attention mechanism to obtain richer semantic features. In another work, a Transformer-based Implicit Latent GAN (TILGAN) model was proposed [42], which combines a transformer autoencoder and GAN in the latent space with a novel design and distribution matching based on the Kullback-Leibler (KL) divergence.

To generate more relevant answers, some researchers used a hybrid models, including the generative and retrieval-based methods. For example, Zhang et al. [43] fed the retrieved answers from the retrieval-based model to a generative-based model as additional information for the discriminator and generator models. Zhu et al. [44] also used the N-best retrieved answers as evidence for calculating the reward for the generator model. Zhang et al. [45] employed the answer generated by retrieval-based models as additional information for training the generative-based models and placed a filter to select the best answer. Furthermore, an ensemble-based deep reinforcement learning was also developed for generative Chatbots by Cuayáhuitl et al. [46].

In short, recent success of ChatGPT has also encouraged researchers to discover further possibilities for chatbot applications[47]. However, despite technological advancements, there are still numerous challenges that must be addressed to create chatbots for accurately depicting human conversation and its features in specific domain. While the generative-based Chatbots are able to generate more relevant answers, these models face some challenges, such as slow convergence due to their high variance and low processing speed especially in large sentences. To overcome the challenge of sequential nature as well as the accuracy improvement in these models, in this paper, we propose a novel architecture based on the cWGAN using transformer model, processing parallelly during the training phase.

### 3. Proposed model

A vanilla GAN model consists of a generator (G) and a discriminator (D). Considering the scope of this paper, the GAN input is a question in the form of sequence x. The discriminator learns to maximize score $D(x, y^*)$ and minimize score $D(x, \hat{y})$, while the generator learns to generate answer $\hat{y}$ to maximize $D(x, \hat{y})$ as expressed in Eq. (1):

$$minmax \, E[logD(x, y^*)] + E[\log(1 - D(x, \hat{y}))] \quad (1)$$

where $(x, y^*)$ is the joint probability distribution of (x, y). $x \sim P_R(x)$ denotes the probability distribution of x from training data. As the GAN models may never converge and have a problem of mode collapses, different variants of the GAN model have been suggested by researchers[48]. One of these variants is the Wasserstein GAN (WGAN), which is an extension of the GAN that seeks an alternate way of training the generator model to better approximate the distribution of data observed in a given training dataset. Instead of using a discriminator to classify or predict the probability of generated data as real or fake, WGAN changes or replaces the discriminator model with a critic that scores the reality or fakeness of a given data. The goal is to minimize the distance between the data distribution in the training dataset and the generated examples. This method can promote stable training while working with gradients. Considering the superiority of the WGAN compared to the GAN model, we propose a novel architecture based on cWGAN and transformer model for generating answers in Chatbot. The intuition behind using WGAN in the proposed method is that there is no Sigmoid activation function to limit the values to 0 or 1 corresponding to real or fake. Also, the training process is more stable when a hybrid of WGAN with the transformer is used. As shown in Fig. 2, the proposed architecture consists of two modules: generator and discriminator which are connected in a single network. The reason for choosing WGAN in the proposed method and hybrid it with the transformer is to increase stability in the training process.

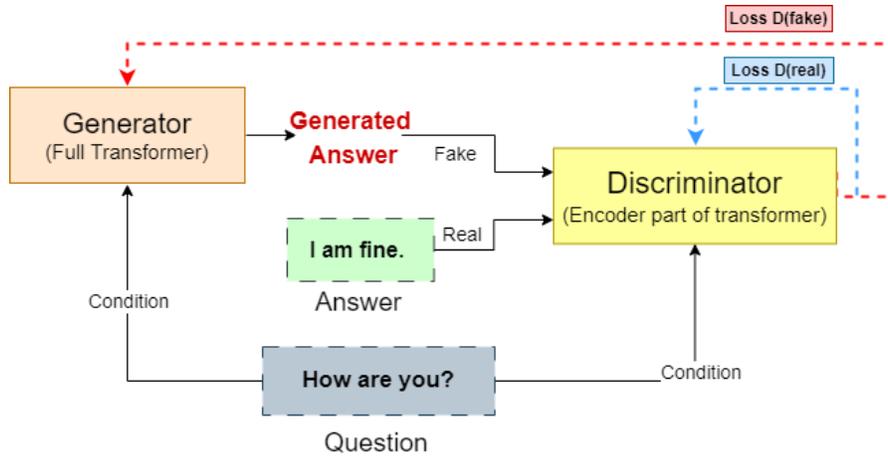

Fig. 2. General architecture of the proposed approach.

### 3.1. Generator modules

Generator is a full transformer model that generates fake answers in test phase. The architecture of the Generator in training phase is illustrated in Fig. 3. The transformer architecture of this figure is adapted from the [20].

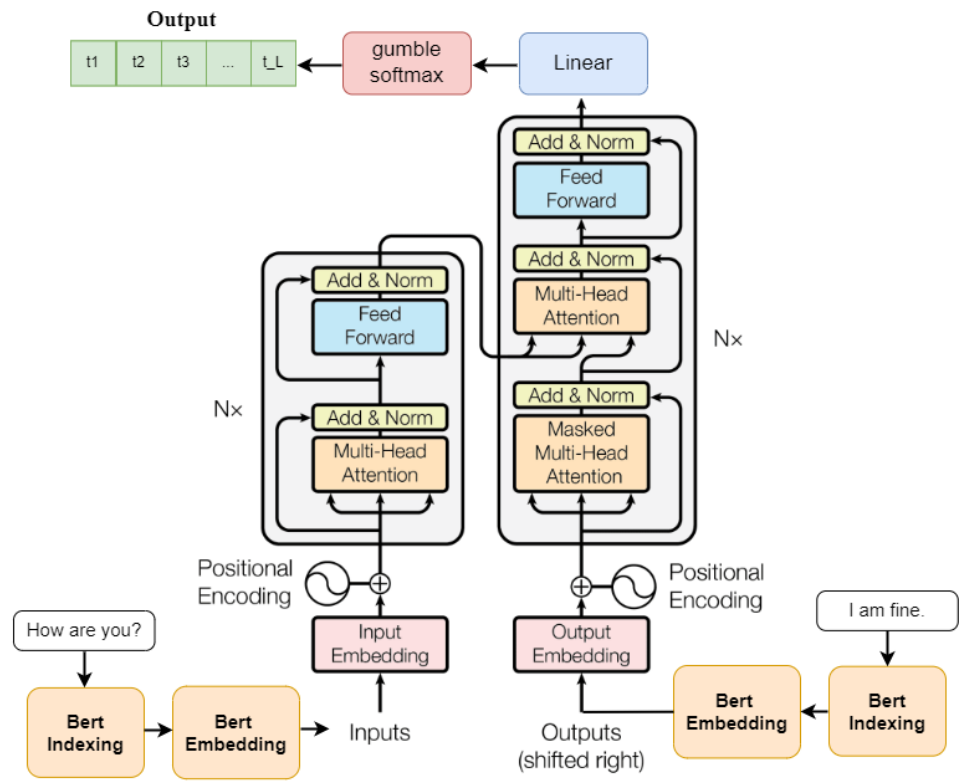

Fig. 3. Architecture of the proposed Generator in training phase.

In the training phase, the encoder and decoder input embedding need to be prepared. Therefore, for the encoder input, the real questions are tokenized and embedded by the pretrained BERT model, including 12 layers, 768 features, and 12 heads. To increase the training speed, a linear model was adopted to reduce the dimensionality of the features to 64, the functionality of this linear model is shown in Eq. (2):

$$y = f(\sum_{i=0}^{n} w_i x_i + \theta) \quad (2)$$

Subsequently, the output of the linear model is concatenated with the position encoding, preserving the word's positions in the sentence. Positional encoding is a matrix, which gives context based on the word position in a sentence as Eq. (3) and Eq. (4):

$$PE_{(pos,2i)} = \sin(pos/10000^{2i/d_{model}}) \quad (3)$$

$$PE_{(pos,2i+1)} = \cos(pos/10000^{2i/d_{model}}) \quad (4)$$

where "pos" refers to the position of the "word" in the sequence; while "d" means the size of the word/token embedding. "i" refers to each of dimensions of the embedding. "d" is fixed, while "pos" and "i" vary.

The same process with some variations is repeated for the decoder. In the decoder, instead of a question, we have a real answer in the input of the decoder. The transformer works slightly different during training and inference phase. During inference, only a question is presented as input sequence. There is not any real answer as target sequence that could be passed to the decoder. Since, decoder aims to generate an answer $\hat{y}$ as close as possible to the real answer, the output is generated in a loop and fed the output sequence from the previous timestep to the decoder in the next timestep till the end token.

In Generator, a full transformer with N = 8 identical layers and 16 heads is used. Generator module is first trained separately by Maximum Likelihood Estimation (MLE) to increase the convergence probability. Then, the model is fine-tuned by adversarial network to learn the answers distribution. The linear transformation and Gumbel SoftMax function are utilized to convert the decoder output to the predicted next-token probabilities. The Gumbel-SoftMax distribution is a continuous distribution capable of approximating samples from a categorical distribution and providing a hard output by using an argmax function. So that, the output of decoder will be an index vector with L length, which is given as the input of the discriminator. In adversarial phase, the generator is updated upon obtaining the discriminator model. In GAN model, this is achieved by gradient likelihood ratios of objective function which can be derived by Eq. (5):

$$L_G = \nabla_{\theta_G} \frac{1}{m} \sum_{i=1}^{m} \log[D(G(Q_i))] \quad (5)$$

where $L_G$ shows the loss function of generator, m is the number of generated sequences while Q denotes the question regarded as condition data. As discussed before, we employ the WGAN, aiming to overcome the challenges of GAN model and increase stability. In this way, the objective function of WGAN, $L_{wG}$, can be calculated by Eq. (6):

$$L_{wG} = \nabla_{\theta_G} \frac{1}{m} \sum_{i=1}^{m} f(G(Q_i)) \quad (6)$$

where f has to be a 1-Lipschitz function.

**3.2. Discriminator modules**

Discriminator includes only encoder part of transformer followed by a classifier. It provides the probability of real or fake answers at the time step of t and trains k epochs. In GAN model, discriminator is updated by gradient likelihood ratios of objective function as expressed by Eq. (7):

$$L_D = \nabla_{\theta_G} \frac{1}{m} \sum_{i=1}^{m} -\log[D(A_i, Q_i)] - \log[1 - D(G(Q_i))] \quad (7)$$

where $L_D$ is loss function of discriminator, A is the real answer, and Q represents the question that is considered as the condition data. In the WGAN model, the discriminator is considered as a critic model; $L_C$ can be calculated by Eq. (8):

$$L_C = \nabla_{\theta_G} \frac{1}{m} \sum_{i=1}^{m} f(q_i) - f(G(q\_i)) \quad (8)$$

where f has to be a 1-Lipschitz function.

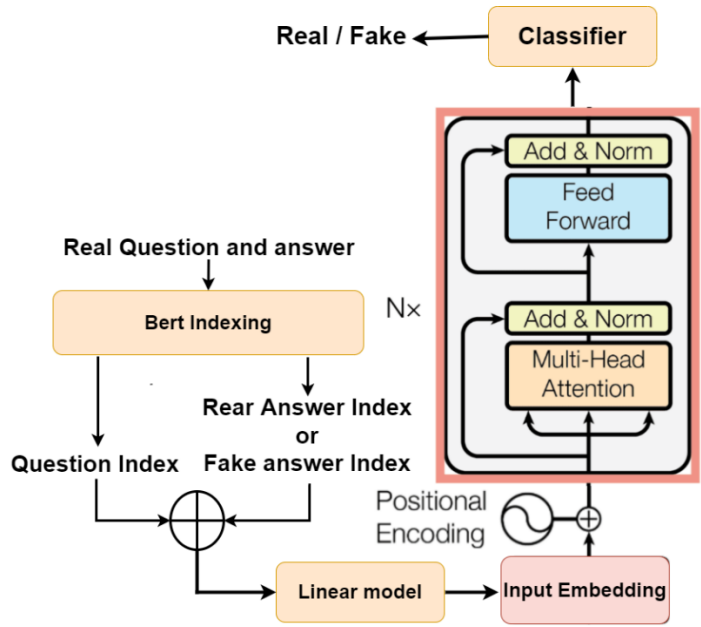
Fig. 4. Architecture of the Discriminator in the proposed model.

As we show in Fig. 4, in each iteration step, the discriminator is trained once with fake pair data and once with real pair data. Real question and answer pairs are tokenized and indexed by the pretrained BERT model. We cannot use BERT for embedding due to the computational complexity of the graph data. Therefore, the index matrix of fake answer is concatenated with index matrix of question as fake pairs. Afterward, a linear model is adopted to reduce the dimensionality of the input matrix and sentence embedding. Then, the output of linear model is concatenated with position encoding matrix to serve as the input matrix for the first layer of the encoder. The output of the last layer of the encoder is fed to a classifier which is a linear network. This network has two scores outputs: the reality or fakeness of a given answer. In the inference phase, we only have generator module while the discriminator model is removed. The decoder architecture of this Fig. 4 is adapted from the [20].

## 4. RESULTS

To demonstrate the performance of the generative Chatbot model, this section presents the details of the dataset and the results in comparison with other methods. First, the implementation details are explained. Then, two datasets used for the evaluation, are briefly introduced, followed by the used evaluation metrics. Finally, the results of the proposed architecture are compared with the state-of-the-art models.

### 4.1. Implementation details

Evaluations were carried out on a Core (TM) i5-12600K with 128GB RAM in Microsoft Windows 11 operating system and Python software with NVIDIA GeForce RTX 3090. The PyTorch library was used to implement the model. The implementation parameters are listed in Table 1.

Table 1. Details of the parameters used in proposed architecture

| Parameters | value | Parameters | value |
|---|---|---|---|
| Learning rate | 0.00005 | Number of layers | 8 |
| Batch size | 64 | Dataset split ration for test data | 20% |
| Epoch numbers | 400 | Sentence Max Length | 30 |
| Processing way | GPU | Number of Heads | 16 |
| Dropout | 0.5 | BERT features size | 768 |

### 4.2. Datasets

Two datasets, Cornell Movie Dialogs Corpus and Chit-Chat, are employed to evaluate the proposed model. Cornell Dataset contains a large metadata-rich collection of fictional conversations extracted from raw movie scripts. This dataset includes 220,579 conversational exchanges between 10,292 pairs of movie characters from 617 movies and a total of 304,713 utterances. The Chit-Chat dataset also encompasses 7,168 conversations from 258,145 utterances and 1,315 unique participants. This dataset was from the Chit-Chat Challenge of the BYU Perception, Control, and Cognition Laboratory.

### 4.3. Evaluation metrics

The proposed model is evaluated using BiLingual Evaluation Understudy (BLEU), Recall-Oriented Understudy for Gisting Evaluation (ROUGE-L), F-measure and Metric for Evaluation of Translation with Explicit ORdering (Meteor). BLEU was originally developed for machine translation evaluation. In this measure, the degree of overlap between the generated sentence and the ground truth sentence is obtained based on n-gram. Unlike BLEU which focuses on precision, ROUGE-L is concentrated on the recall as it needs to calculate the similarity of the generated sentence and the ground truth. F-measure is a valuable metric for evaluating the performance of classification algorithms, which can be defined as a compromise between recall and precision. METEOR is a common evaluation measure that is widely used to measure the similarity between sentences. This metric provides the closest results to human judgments.

### 4.4. Experimental results

Here, the effectiveness of the proposed architecture is presented on the Cornell Movie Dialogs Corpus and Chit-Chat datasets using the BLEU4, ROUGE-L, F-measure and Meteor metrics (See Table 2). As the results of this table confirm, the proposed model has a better performance on the Chit-Chat dataset. This comes from this point that the Chit-Chat dataset is based on chat and conversation while the Cornell dataset relies on movie dialogues.

Table 2. Performance of proposed architecture in terms of different metrics

| Dataset | BLEU4 | ROUGE-L | F-measure | Meteor |
|---|---|---|---|---|
| Cornell | 0.71 | 0.818 | 0.622 | 0.52 |
| Chit-Chat | 0.96 | 0.965 | 0.989 | 0.87 |

Several state-of-the-art approaches, including S2S+Attention[31], BERT[31], Retrieval-augmented generation (RAG) [49], Open Domain Response Generation (OPRG) [31], BILSTM and Deep Transfer learning-based Text Generation (DTGEN) [36], are used to compare to the proposed model with ROUGE-L metric on Cornell dataset (Table 3). According to Table 3, the proposed model outperforms all models due to the extracting richer features by transformer as well as learning data distribution by GAN.

Table 3. The comparison results of the proposed model with the different methods on Cornell dataset using ROUGE-L metric.

| Method | Approach | ROUGE-L | Year |
|---|---|---|---|
| S2S+Attention | Seq2Seq | 0.337 | - |
| BERT | BERT +LSTM | 0.399 | - |
| RAG | Retrieval model+ BERT | 0.427 | 2020 |
| OPRG | Retrieval model+ Transformer | 0.452 | 2022 |
| BILSTM | Seq2Seq | 0.74 | - |
| DTGEN | Transformer + Seq2Seq | 0.815 | 2022 |
| **Proposed** | GAN+ Transformer | **0.818** | - |

Table 4 shows the comparison results of the proposed model with the MLE[22], SeqGAN[21], StepGAN[22] and CaiRE[39] in Chit-Chat dataset using the BLEU4 metrics. The approaches used for comparison employ GAN model for generating sequence data like text. According to Table 4, the proposed model outperforms all approaches using the BLEU-4 metric in the Chit-Chat dataset due to using an extended hybrid of transformer model and cWGAN methods. Compared to the state-of-art methods, the proposed model takes the advantage of data distribution learning in answer generation to modify all evaluation metrics. According to the experimental results, the proposed model can generate more accurate and semantically relevant answers for the Chatbot dialogue.

Table 4. The comparison results of the proposed model with the different methods on Chit-Chat dataset using BLEU-4 metric.

| Method | Approach | BLEU-4 | Year |
|---|---|---|---|
| MLE | Seq2Seq | 0.25 | - |
| SeqGAN | GAN +MC | 0.26 | 2017 |
| StepGAN | GAN +QN | 0.23 | 2019 |
| CAiRE | Transformer | 0.70 | 2020 |
| **Proposed** | **GAN+ Transformer** | **0.96** | - |

For investigating the generated answers of the proposed Chatbot, various questions from different domains were asked from the proposed Chatbot. Table 5 shows the results of the Chatbot trained using Cornell dataset in two areas of greeting and general questions and conversations. In addition, Table 6 shows the results of the proposed Chatbot trained on the Chit-Chat dataset in two areas of greeting and general questions and conversation.

Table 5. Some examples of the generated answers using the proposed Chatbot on Cornell dataset.

| **Greeting and General question** | | |
|---|---|---|
| **User input** | **Real answer** | **Chatbot answer** |
| hello Mr. parker how are you? | hello Jo thanks you | hello Ju thank you |
| I heard that was you good? | well, it was nice seeing you | well, it was good seeing you |
| how much your goanna takes? | I do not know how much do you want | I do not know how much do you want |
| hey | hey what are you doing right now | hi what are you doing right now |
| how many girlfriends did you have? | I do not know exactly | I do not know truly |
| feeling better? | I just cannot believe it | I actually not believe it |
| **General Conversation** | | |
| **User input** | **Real answer** | **Chatbot answer** |
| is this for like real? | unfortunately, yes | respect yes |
| how do you know that? | she told me, he was in jail | he told me, maybe they have in jail |
| I tried talking to her at the mine it did not work | then try again | then go again |
| I think you are lying that has that is what kind of problem | what are you saying | what are you saying |
| oh, I put a few games on for your daughter I hope you do not mind | of course, not | sure, course not |
| I am not playing a game now | yes, you are playing word games | yes, you are playing cook games |
| there is no other way? | but there is another entrance to this place | there is another answer to this place |
| I have changed my mind | you cannot change your mind fully | you should not change your mind fully |
| hey ben there were a couple of guys looking for you | what did they look like | what did they look like |

Table 6. Some examples of the generated answers using the proposed Chatbot on Chit-Chat dataset.

| **Greeting and General question** | | |
|---|---|---|
| **User input** | **Real answer** | **Chatbot answer** |
| can you hear me? | Hello there | Hello there |
| how are you today? | I am great thanks for asking me | I am great thanks for asking |
| can you talk with me? | sure, ask me a question | sure, ask me a question |
| what age are you? | I am a bot so I do not have an age | I am a bot so I do not have an age |
| I am leaving now | goodbye | goodbye |
| **General Conversation** | | |
| **User input** | **Real answer** | **Chatbot answer** |
| what kind of thing can you respond to? | I am here to help answer your questions | I am here to help answer your questions |
| how many sisters do you have? | I do not have a family the same way humans do | I do not have a family the same way humans do |
| do you have a gender identity? | since I am digital, I do not actually have a gender | since I am digital I do not actually have a gender |
| do you think you are the most intelligent | we think in very different ways but it has it is safe to say you are smarter | we think in very different ways but it has it is safe to say you are smarter |

## 5. DISCUSSION

The main challenge of most recent Chatbot approaches is their sequential nature, which is led to less accurate results and performance. Proposed model resolves this challenge by using a hybrid of cWGAN and transformer model. According to the experimental results, our model can generate more accurate and semantically relevant answers for the Chatbot dialogue. However, the proposed model benefits from the answer distribution learning, the generated answers have little diversity. This is due to the nature of the dataset used in the training phase. Both datasets lack of multiple answers for each question. Here, we discuss the proposed model from two perspectives as follows:

- **Analysis of generator pretraining:** Before the adversarial training of the proposed model, the generator model is trained separately by a transformer model. The point is that the pre-training of the generator model has a direct impact on the performance of the whole proposed model. In the best effort, 200 cycles are used to train the transformer as a pre-trained model of the generator. After that, the proposed model is trained by adversarial learning in 400 cycles. According to Table 7, the model that is trained only with transformer without including the adversarial learning has the lowest performance. We considered this model as a pretrained model to start training in adversarial phase. The results show that it improves the performance of the Chatbot in all metrics of the Cornell dataset.

Table 7. Performance of the proposed model with different configurations on Cornell dataset.

| Type of model | BLEU4 | ROUGE-L | F-measure | Meteor |
|---|---|---|---|---|
| Only generator pretraining | 0.677 | 0.789 | 0.531 | 0.481 |
| Only Adversarial | 0.689 | 0.801 | 0.573 | 0.492 |
| **Combine generator pretraining with Adversarial** | **0.71** | **0.818** | **0.62** | **0.521** |

- **Analysis of loss function:** Loss function plays a key role in the performance of deep models. In this part, the loss functions of generator and discriminator in adversarial phase are considered and analyzed. In the adversarial training phase, the WGAN is used instead of GAN model. WGAN uses a new cost function using Wasserstein distance which has a smoother gradient and trains regardless of the implementation of the generator. Wasserstein distance is calculated by Eq. (10):

$$W(P_{real}, P_{fake}) = \sup E_{x \sim P_{real}}[f(x)] - E_{x \sim P_{fake}}[f(x)] \quad (10)$$

where sup is the least upper bound, f shows a Lipschitz function, and x denotes a real or fake answer. To calculate the Wasserstein distance, we just need to find a Lipschitz function. We can build a deep network to learn it. This network is similar to the discriminator, but it has no Sigmoid function and outputs a scalar score rather than the probability. This score can be interpreted as the realness of the input data and considered as critic. The Wasserstein loss function can be summarized for generator and critic (discriminator) in proposed model in Eq. (11) and Eq. (12), respectively:

$$GLoss = -[avg\ critic\ score\ on\ fake\ answer] \quad (11)$$

$$CLoss = [avg\ critic\ score\ on\ real\ answer] - [avg\ critic\ score\ on\ fake\ answer] \quad (12)$$

According to Fig. 5, in Cornell dataset, the loss function of the generator is first low due to having a generator pretraining model. After a few epochs, the loss function increases followed by a smooth decrease to less than its initial value. This represents the learning of data distribution in the adversarial learning. Moreover, the discriminator has a low loss function at the beginning. Upon learning the data distribution by the generator and generating better fake data, the value of the discriminator loss function smoothly increases. Eventually, both generator and discriminator loss functions are converged.

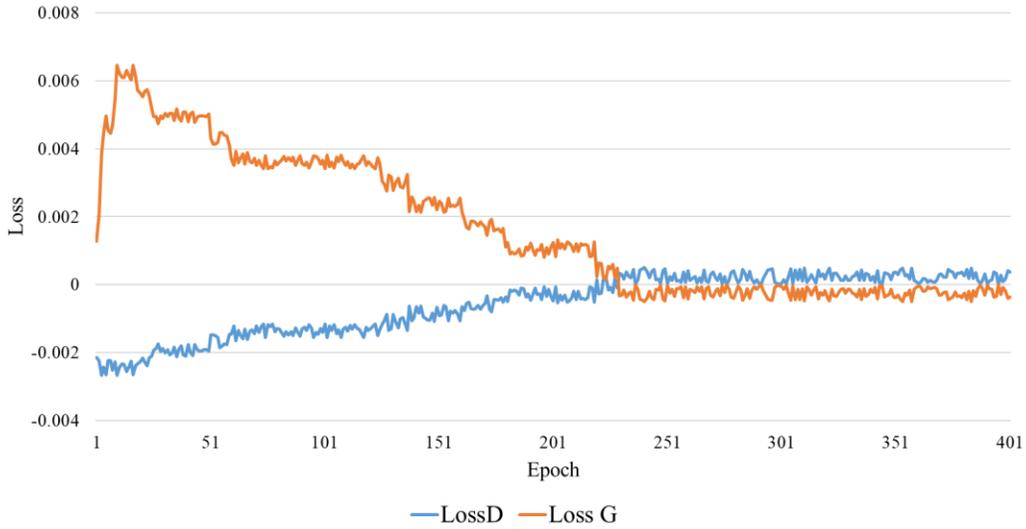

Fig. 5. Generator (LossG) and discriminator (LossD) loss in Cornell dataset.

According to Fig. 6, in Chit-Chat dataset, the loss function in the generator is initialized with a positive value close to zero, due to its generator pretraining model. After several epochs, the loss function increases followed by a slow degradation to close-zero values. This value is less than the initial value, which indicates the improvement of chatbot efficiency with the addition of the adversarial model. Furthermore, the discriminator initially has a loss function close to zero, which gradually increases to higher values as the generator learns the data distribution. This indicates better fake answer generation by the generator.

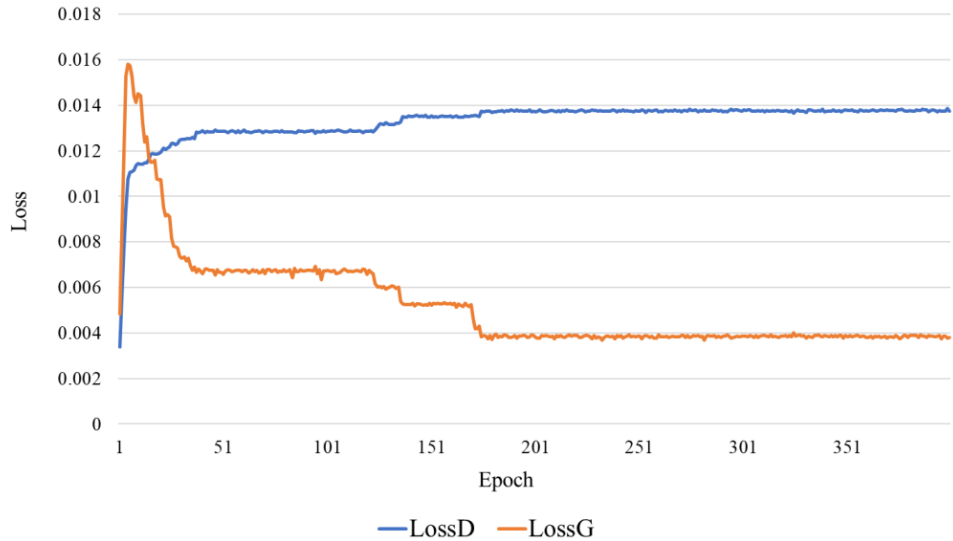

Fig. 6. Generator (LossG) and discriminator (LossD) loss in Chit-Chat dataset.

## 6. CONCLUSION AND FUTURE TREND

In this paper, we proposed a novel model using the combination of the cWGAN and transformer model. The proposed model consists of two networks: Generator and Discriminator. Generator is a full Transformer model and Discriminator includes only the encoder part of a transformer model followed by a classifier. We evaluated the proposed model on two datasets using different evaluation metrics. The results confirmed the superiority of the proposed model compared to state-of-the-art approaches according to BLEU4, ROUGE-L, F-measure and Meteor metrics. Relying on the cWGAN capabilities as well as the transformer model, the proposed model generates accurate, semantically relevant, and human-like answers. Future works can add reinforcement leaning to increase semantic relations between question and answer in various domains.

## DECLARATIONS
**Funding**
This research did not receive any specific grant from funding agencies in the public, commercial, or not-for-profit sectors.

**Data Availability**
The datasets analyzed during the current study are available on these web pages:

Cornell_Movie-Dialogs_Corpus: https://www.cs.cornell.edu/~cristian/Cornell_Movie-Dialogs_Corpus.html
Chit-Chat dataset: https://pypi.org/project/chitchat-dataset/

**Declaration of competing interest**
The authors certify that they have no conflict of interest.